%% file: main.tex
\documentclass[letterpaper, 10 pt, conference]{ieeeconf}
\IEEEoverridecommandlockouts
\usepackage[table]{xcolor}
\usepackage{algorithm,algcompatible}
\usepackage{array}
\usepackage{graphicx}
\usepackage{wrapfig,lipsum}
\usepackage{amsmath}
\usepackage{amssymb}
\makeatletter
\let\NAT@parse\undefined
\makeatother
\usepackage{hyperref}
\usepackage{gensymb}
\usepackage{booktabs}
\usepackage{multirow}
\usepackage{nicematrix}
\usepackage{diagbox}
\usepackage{slashbox}
\usepackage{tikz}
\usepackage{caption}
\DeclareCaptionFormat{figure}
{#1.~#3
}
\DeclareCaptionFormat{table}
{#1\\\textsc{#3}
}
\title{\LARGE\bf Differentiable Parsing and Visual Grounding of Natural Language Instructions for Object Placement
\thanks{The authors are from National University of Singapore. Emails: \{ziruiz, leews, dyhsu\}@comp.nus.edu.sg.
   \newupdate{This research is supported in part by the National Research Foundation (NRF), Singapore and DSO National Laboratories under the AI Singapore Program (AISG Award No. AISG2-RP-2020-016) and the Agency of Science, Technology and Research, Singapore, under the National Robotics Program (Grant No. 192 25 00054).}}
\thanks{The data, code and appendix are available at \href{https://bit.ly/ParaGonProj}{\color{blue}https://bit.ly/ParaGonProj}.}}

\newcommand{\bfx}{\mathbf{x}}
\newcommand{\bfy}{\mathbf{y}}
\newcommand{\obs}{\mathbf{o}}

\newcommand{\lang}{\mathcal{L}}

\newcolumntype{P}[1]{>{\centering\arraybackslash}p{#1}}

\algnewcommand\INPUT{\item[\textbf{Input:}]}%
\algnewcommand\OUTPUT{\item[\textbf{Output:}]}%
\algnewcommand\PARAM{\item[\textbf{Layer params:}]}%

\newif\ifreview

\ifreview
\newcommand{\newupdate}[1]{{\color{blue}#1}}
\else
\newcommand{\newupdate}[1]{#1}
\fi

\author{
    Zirui Zhao, Wee Sun Lee, and David Hsu
}

\begin{document}
\maketitle


\input{0_abstract.tex}

\input{1_intro}

\input{2_related_work}

\input{3_method}
\input{4_exp}
\input{5_conclustion}





\bibliographystyle{IEEEtran}
\bibliography{ref}  
\end{document}

%% file: 0_abstract.tex
\begin{abstract}
We present a new method, PARsing And visual GrOuNding~(\textsc{ParaGon}), for grounding natural language in object placement tasks. Natural language generally describes objects and spatial relations with \emph{compositionality} and \emph{ambiguity}, two major obstacles to effective language grounding. For compositionality, \textsc{ParaGon} parses a language instruction into an object-centric graph representation to ground objects individually. For ambiguity, \textsc{ParaGon} uses a novel particle-based graph neural network to reason about object placements with uncertainty. Essentially, \textsc{ParaGon} integrates a parsing algorithm into a probabilistic, data-driven learning framework. It is fully differentiable and trained end-to-end from data for robustness against complex, ambiguous language input.

\end{abstract}

%% file: 1_intro.tex
\section{Introduction}

Robot tasks, such as navigation, manipulation, assembly, \ldots,  often involve spatial relations among objects. 
To carry out tasks instructed by humans, robots must understand natural language instructions about objects and their spatial relations.
This work focuses specifically on the object placement task with language instructions.  Humans provide verbal instructions to robots to pick up an object and place it in a specific location. The robot must generate object placements based on both language description and visual observation. However, the language expressions about spatial relations are generally ambiguous and compositional, two major obstacles to effective language grounding.


Two types of ambiguity, \textit{positional ambiguity} and \textit{referential ambiguity}, are focused upon in this study. \textit{Positional ambiguity} occurs when humans describe directional relations without specifying exact distances (e.g., to the left side of a plate''). Reference objects are often required to describe spatial relations. When placing an object next to a reference object without specifying its distances, it is hard to link the reference expressions to the referred objects to learn visual grounding. \textit{Referential ambiguity} arises when descriptions of objects are ambiguous, leading to a reference expression being grounded to multiple semantically identical objects and resulting in a multimodal distribution of correct placement.

\begin{figure}[t]
    \centering
    \includegraphics[width=\columnwidth]{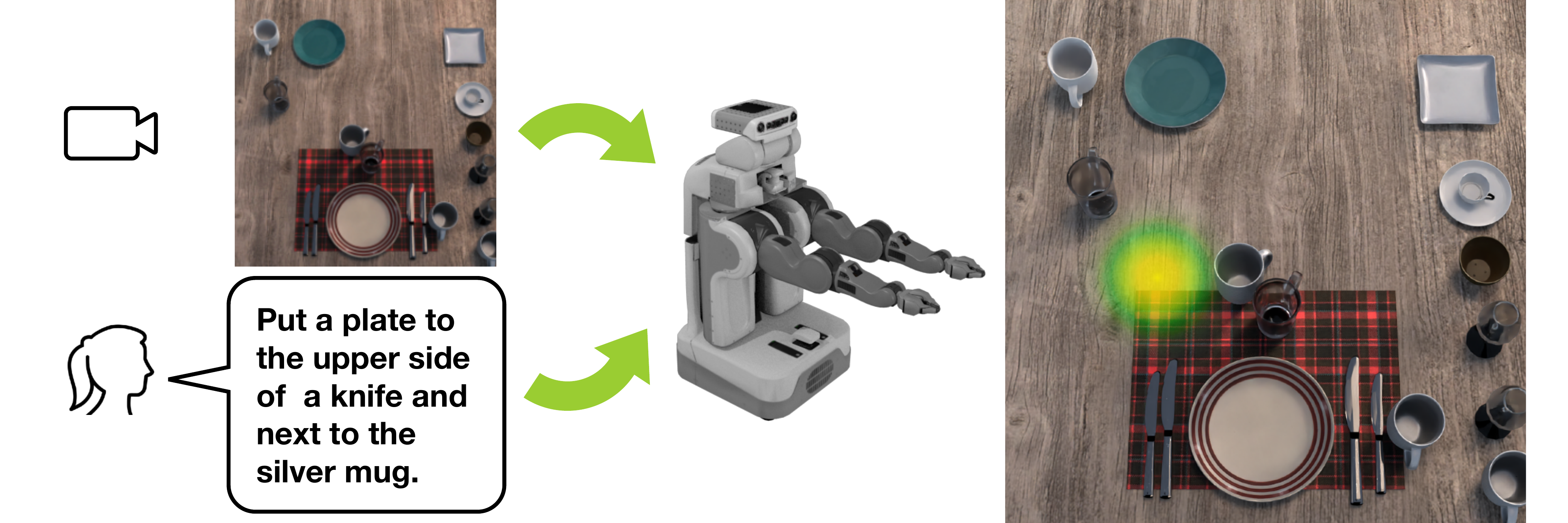}
    \caption{\textsc{ParaGon} takes as input a language instruction and an image of the task environment. It outputs candidate placements for a target object.   
    The presence of multiple semantically identical objects (e.g., ``silver mug'') and omitted distance information cause difficulty for placement generation, and the compositional instructions increase the data required for learning.}\label{fig:demo}
\end{figure}

The compositional structure of language-described spatial relations stems from the visual scene's and natural language's compositional nature. A complex scene includes multiple basic objects, and to describe the desired state of a complex scene, one can compose many simple sentences for referents and their relations to form a complex language sentence (e.g., the instruction in Fig~\ref{fig:demo}). This characteristic increases the data required for learning compositional language instructions.

To address the issues, we propose \textsc{ParaGon}, a PARsing And visual GrOuNding method for language-conditioned object placement.
The core idea of \textsc{ParaGon} is to parse human language into object-centric relations for visual grounding and placement generation, and encode those procedures in neural networks for end-to-end training. The parsing module of \textsc{ParaGon} decomposes compositional instructions into object-centric relations, enabling the grounding of objects separately without compositionality issues. The relations can be composed and encoded in graph neural networks (GNN) for placement generation. This GNN uses particle-based message-passing to model the uncertainty caused by ambiguous instructions. All the modules are encoded into neural networks and connected for end-to-end training, avoiding the need for individual training module labeling.

\textsc{ParaGon} essentially integrates parsing-based methods into a probabilistic, data-driven framework, exhibiting robustness from its data-driven property, as well as generalizability and data efficiency due to its parsing-based nature. It also adapts to the uncertainty of ambiguous instructions using particle-based probabilistic techniques. The experiments demonstrate that \textsc{ParaGon} outperforms the state-of-the-art method in language-conditioned object placement tasks in the presence of ambiguity and compositionality.

%% file: 2_related_work.tex
\section{Related Work}
\newupdate{Grounding human language in robot instruction-following has been studied in the recent decades~\cite{shridhar2018interactive, shridhar2020ingress, mees2020learning, mees2020composing, liu2021structformer, kartmann2021semantic, shridhar2021cliport, stepputtis2020language, lynch2021language}.} Our research focuses on object placement instructed by human language. In contrast to picking~\cite{shridhar2018interactive, shridhar2020ingress} that needs only a discriminative model to ground objects from reference expressions, placing~\cite{mees2020learning, mees2020composing, liu2021structformer, kartmann2021semantic, shridhar2021cliport} requires a generative model conditioned on the relational constraints of object placement. It requires capturing complex relations between objects in natural language, grounding reference expression of objects, and generating placement that satisfies the relational constraints in the instructions. 

Parsing-based methods for robot instruction-following~\cite{mees2020learning, mees2020composing, kartmann2021semantic, raman2013sorry, boteanu2016model, howard2014natural, boteanu2017robot,kress2018synthesis} parse natural language into formal representations using hand-crafted rules and grammatical structures. Those hand-crafted rules are generalizable but not robust to noisy language~\cite{tellex2020robots}. Among these studies, those focusing on placing~\cite{mees2020learning, mees2020composing, kartmann2021semantic} lack a decomposition mechanism for compositional instructions and assume perfect object grounding without referential ambiguity.  
Recently,~\cite{stepputtis2020language, lynch2021language, shridhar2021cliport} used sentence embeddings to learn a language-condition policy for robot instruction following, which are not data-efficient and hard to generalize to unseen compositional instructions. We follow~\cite{karkus2019differentiable} to integrate parsing-based modules into a data-driven framework. It is robust, data-efficient, and generalizable for learning compositional instructions. We also use probabilistic techniques to adapt to the uncertainty of ambiguous instructions. 

\textsc{ParaGon} has a GNN for relational reasoning and placement generation, which encodes a mean-field inference algorithm similar to~\cite{dai2016discriminative}. Moreover, our GNN uses particles for message passing to capture complex and multimodal distribution. 
It approximates a distribution as a set of particles~\cite{dauwels2006particle}, which provides strong expressiveness for complex and multimodal distribution. It was used in robot perception~\cite{gustafsson2010particle, karkus2018particle, zhu2020towards}, recurrent neural networks~\cite{ma2020particle}, and graphical models~\cite{sudderth2003nonparametric,su2020multiplicative}. Our approach employed this idea in GNN for particle-based message passing.

%% file: 3_method.tex
\section{Overview}

\newupdate{We focus on the grounding human language for tabletop object placing tasks.} In this task, scenes are composed of a finite set of 3D objects on a 2D tabletop. Humans give natural language instruction $\ell\in\lang$ to guide the robot to pick an object and put it at the desired position $\bfx^*_\text{tgt}$. The human instruction is denoted as a sequence $\ell = \{\omega_l\}_{1\leq l\leq L}$ where $\omega_l$ is a word, e.g., instruction in Fig.\ref{fig:examples} is $\{\omega_1=\text{put}, \omega_2=\text{a}, \ldots\}$. 
An instruction should contain a target object expression (e.g., ``a plate'') to specify the object to pick and express at least one spatial relation (e.g., ``next to a silver mug'') for placement description. The robot needs to find the target object's placement distribution $p(\bfx^*_\text{tgt}|\ell, \obs)$ conditioned on the language instruction $\ell$ and visual observation $\obs$. 

\begin{figure*}[h]
\centering
\includegraphics[width=.9\textwidth]{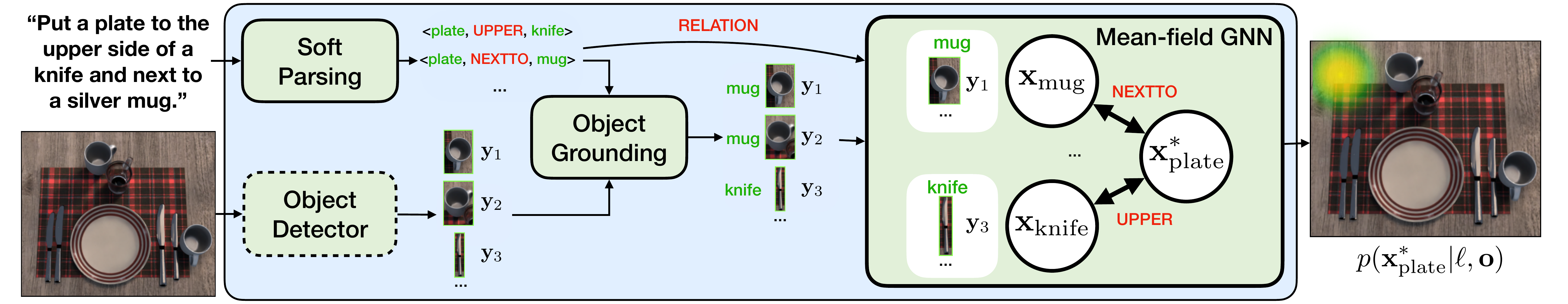}
\caption{An overview of \textsc{ParaGon}. \textsc{ParaGon} uses soft parsing to represent language input as relations between objects. The grounding module then aligns the stated objects to the object-centric regions in the visual scene. A GNN then conducts relational reasoning between grounded objects and outputs placement. \textsc{ParaGon} is fully differentiable and trained end-to-end without labels for parsing and visual grounding.}\label{fig:key}
\end{figure*}



\newupdate{The core idea in our proposed solution, \textsc{ParaGon}, is to leverage object-centric relations from linguistic and visual inputs to perform relational reasoning and placement generation, and encode those procedures in neural networks for end-to-end training.}
The pipeline of \textsc{ParaGon} is in Fig~\ref{fig:key}. 
\textsc{ParaGon} first uses the soft parsing module to convert language inputs ``softly'' into a set of relations, represented as triplets. 
A grounding module then aligns the mentioned objects in triplets with objects in the visual scenes. 
The triplets can form a graph by taking the objects as the nodes and relations as the edges. The resulting graph is fed into a GNN for relational reasoning and generating placements. The GNN encodes a mean-field inference algorithm for a conditional random field depicting spatial relations in triplets.
\textsc{ParaGon} is trained end-to-end to achieve the best overall performance for object placing without annotating parsing and object-grounding labels.

\section{Soft Parsing}

\begin{figure}[t]
    \includegraphics[width=\columnwidth]{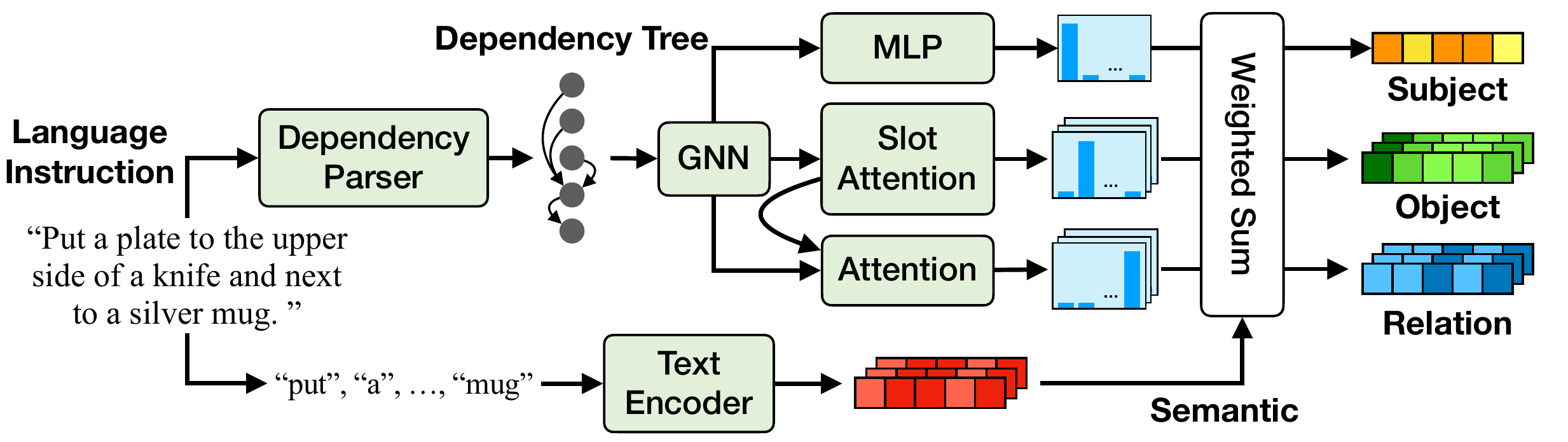}
    \caption{The pipeline of soft parsing module.}
    \label{fig: soft_par} 
\end{figure}

The soft parsing module is to extract spatial relations in complex instructions for accurate placement generation. The pipeline is in Fig~\ref{fig: soft_par}. 
Dependency trees capture the relations between words in natural language, which implicitly indicate the relations between the semantics those words express~\cite{schuster2015generating, wang2018scene}. Thus, we use a data-driven approach to explore the underlying semantic relations in the dependency tree for extracting relations represented as relational triplets. It takes linguistic input and outputs relational triplets, where the triplets' components are represented as embeddings. 

\subsection{Preliminaries}



\subsubsection*{Triplets} A triplet consists of two entities and their relation, representing a binary relation. \newupdate{Triplet provides a formal representation of knowledge expressed in natural language, which is widely applied in scene graph parsing~\cite{schuster2015generating}, relation extraction~\cite{guo2019attention}, and knowledge graph~\cite{wang2017knowledge}.}
\newupdate{The underlying assumption of representing natural language as triplets is that natural language rarely has higher-order relations, as humans mostly use binary relations in natural language~\cite{rubinstein1996certain}.}
For spatial relations, two triplets can represent ternary relations (e.g., ``between A and B'' equals ``the right of A and left of B'' sometimes). As such, it is sufficient to represent instructions as triplets for common object-placing purposes. 

\begin{figure}[t]
    \centering
    \includegraphics[width=0.7\columnwidth]{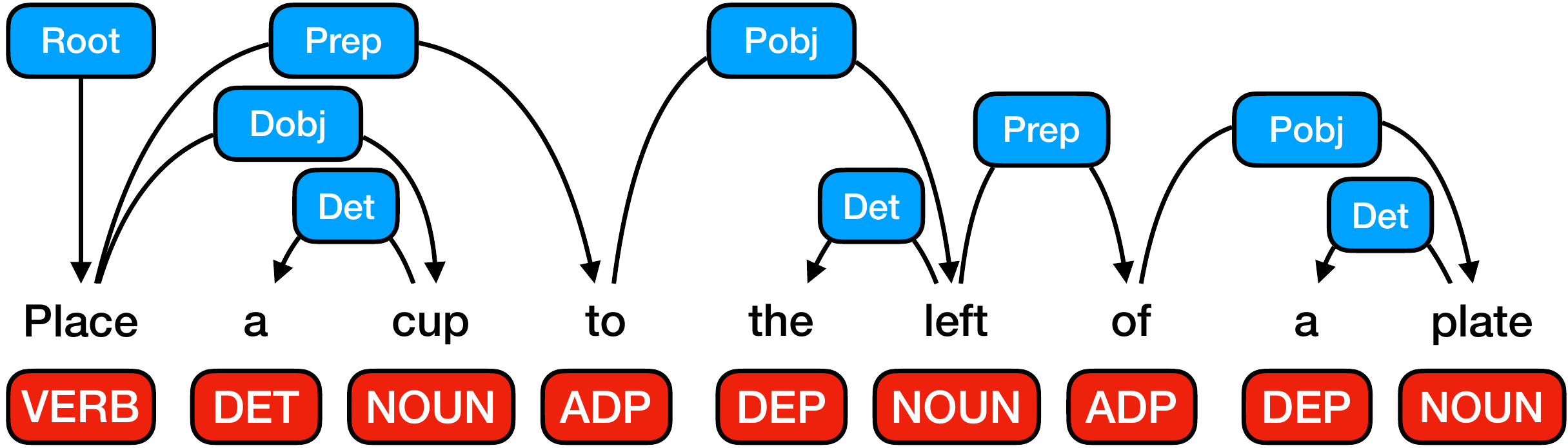}
    \caption{Dependency parsing takes language sequence as the input and outputs a tree data structure with tags. The blue blocks are dependency tags, while the red ones are part-of-speech tags. \newupdate{A part-of-speech tag categorizes words' correspondence with a particular part of speech, depending on the word's definition and context. Dependency tags mark two words relations in grammar, represented as Universal Dependency Relations.}}
    \label{fig: dep tree}
\end{figure}


\subsubsection*{Dependency Tree} A dependency tree (shown in Fig.~\ref{fig: dep tree}) is a universal structure that examines the relationships between words in a phrase to determine its grammatical structure~\cite{kubler2009dependency}. It uses part-of-speech tags to mark each word and dependency tags to mark the relations between two words. \newupdate{A part-of-speech tag~\cite{petrov2011universal} categorizes words' correspondence with a particular part of speech, depending on the word's definition and context, such as in Fig.~\ref{fig: dep tree}, ``cup'' is a ``Noun''. Dependency tags mark two words relations in grammar, represented as Universal Dependency Relations~\cite{mcdonald2013universal}. For example, in Fig.~\ref{fig: dep tree}, the Noun ``cup'' is the ``direct object (Dobj)'' of the Verb ``place''. Those relations are universal.} A proper dependency tree relies on grammatically correct instructions, whereas noisy language sentences may result in imperfect dependency trees. Thus, we use the data-driven method to adapt to imperfect dependency trees. 



\subsection{Method}

To make the parsing differentiable, the soft parsing module ``softens'' the triplet as the attention to the words of linguistic inputs $\{a_{\gamma, l}\}_{1\leq l \leq L}, \sum_la_{\gamma, l}=1, \gamma\in\{\mathrm{subj, obj, rel}\}$. We compute the embeddings of components in triplets by the attention-weighted sum of the word embeddings $\varphi_\gamma = \sum_l a_{\gamma, l}f_\text{CLIP}(\omega_l)$. The word embeddings are evaluated using pre-trained CLIP~\cite{radford2021learning} $f_\text{CLIP}$. \newupdate{As the dependency tree examines the relations between words that could implicitly encode relational information,} we use a GNN~\cite{gilmer2017neural} to operate over the dependency tree from spaCy~\cite{spacy2} for the structural features. In the dependency tree, the part-of-speech tags are node features, and the dependency tags are the edge attributes. Then, we feed the structural features into three modules to evaluate each word's weight to indicate how much this word contributes to the components in a triplet. We use a single layer MLP to compute the attention of subject words $\{a_{\text{subj}, l}\}_{1\leq l\leq L}$ and slot attention module~\cite{locatello2020object} to get $N$ features and attentions $\{a^n_{\text{obj}, l}\}_{1\leq n\leq N, 1\leq l\leq L}$ of object phrases, where $N$ is the total number of triplets. We feed the features of object phrases as the query, and structural features as the key and value of attention, into a multi-head attention layer~\cite{vaswani2017attention} to get the attention for corresponding relational phrases $\{a^n_{\text{rel}, l}\}_{1\leq n\leq N, 1\leq l\leq L}$. We compute the attention-weighted sum of word embeddings from pre-trained language models to get the embeddings of $N$ possible relational triplets, denoted as $\Omega=\{\langle \varphi_\text{subj}, \varphi_\text{rel}^n, \varphi_\text{obj}^n\rangle\}_{n=1}^N$.

\newupdate{Ideally, each triplet should contain information about single objects and their relations. However, we discovered that the features of multiple objects and their relations could be entangled in one triplet. As the disentangled triplets can provide more precise information and be more accurate in recovering the sentence}, we add sentence recovery to guide extracting disentangled triplets. We first compute the embedding of each token using attention:
$\tilde{\varphi}_l=\sum_{n=1}^N\sum_{i\in\{\text{rel}, \text{obj}\}} a_{i,l}^n\varphi^n_i+a_{\text{subj}, l}\varphi_\text{subj}$.
We then feed the embedding of tokens to an LSTM to produce the recovered word embeddings:
$\tilde{\ell}=\{\tilde{\omega}_l\}_{l=1}^L=f_\text{LSTM}(\{\tilde{\varphi}_l\}_{l=1}^L)$. We minimize the L2 loss between generated sequence $l$ and $\tilde{l}$ as an auxiliary task to improve the soft parsing component. 

\section{Visual Grounding}

After relation extraction, we ground the mentioned objects to the visual scenes and reason about their spatial relations for placement generation. We align the visual features of objects and the embeddings of their reference expressions into the same embedding space for object grounding. We then represent relations in conditional random fields and encode them into GNNs for reasoning and placement generation.

\subsection{Object Reference Grounding}

The object reference grounding module aligns the object and subject phrase embeddings into the objects in the visual scene. 
Given the visual inputs, we use a pre-trained object detector (Mask-RCNN~\cite{he2017maskrcnn} or SPACE~\cite{Lin2020SPACE}) to get object bounding boxes $\{\bfy_m\}_{m=1}^M$ and encode their visual features $\{\mathbf{z}_m\}_{m=1}^M$ using a pre-trained CLIP~\cite{radford2021learning}. We then project the object visual features $\obs_m$ and linguistic features in triplets $\varphi_i,~\varphi_i\in\{\varphi_\text{subj}\}\cup\{\varphi^n_\text{obj}\}_{n=1}^N$ into the same feature space via learnable projecting matrices ${\Phi}, {\Psi}$. We evaluate the cosine similarity $d_\text{cos}(\cdot, \cdot)$ with a learnable scaling factor $\alpha$ between the visual and text feature to get the grounding belief:
$b^i_m=\mathrm{Softmax}_m(\alpha\cdot d_\mathrm{cos}(\Phi\mathbf{z}_m^\top, \Psi\varphi_i^\top))$.
As such, the grounding results of object $i$ in the triplets is a set of object-centric regions with belief $\{(b^i_m, \bfy_m)\}_{m=1}^M$. 


\subsection{Relational Reasoning for Placement Generation}

\subsubsection{Spatial relations in Conditional Random Field} 
The triplets essentially build up a relational graph $\mathcal{G}=(\mathcal{V}, \mathcal{E})$ with objects as the nodes $\mathcal{V}$ and their spatial relations as the edges $\mathcal{E}$. 
We use the conditional random field (CRF) specified using $\mathcal{G}$ to represent the relations between positions of the context and target objects in instruction.
The variable for the vertices $\mathbf{X}=\{\bfx_i|\bfx_i\in\mathbb{R}^2\}_{i\in\mathcal{V}}$ denotes the grounded position of each context object and the placement position of the target object. 
We formulate the CRF as: $p(\mathbf{X}|\obs,\Omega)\propto\prod_{(i,j)\in\mathcal{E}} \psi_{ij}(\bfx_i, \bfx_j | r_{ij})\prod_{i\in\mathcal{V}/\{\text{subj}\}}\phi_i(\bfx_i, \obs). $
$\psi_{ij}$ describes the spatial relations between two objects based on the spatial relation $r_{ij}\in\{\varphi^n_\text{rel}\}_{n=1}^N$. 
$\phi_i$ denotes the probability of the context object's position conditioned on observation, except for the target object (subj phrase). It is because the grounded position of the target object is only useful for picking rather than placing. 
Mean-field variational inference approximates the CRF into a mean-field $p(\mathbf{X}|\obs,\Omega)\approx q(\mathbf{X})=\prod_{i\in\mathcal{V}}q_i(\bfx_i)$ and optimizes the reversed KL divergence $\mathrm{KL}(q||p)$ to converge to the multimodal distribution~\cite{bishop2006pattern}. The equation of Mean-field variational inference is: 
$\log q^t_i(\bfx_i)=c_i + \log\phi_i(\bfx_i, \obs)+\sum_{j\in{N}(i)}\int_\mathcal{X}q^{t-1}_j(\bfx_j)\log\psi_{ij}(\bfx_i, \bfx_j|r_{ij})\phi_j(\bfx_j, \obs)d\bfx_j = c_i + m^t_{\text{obs},i}(\bfx_i) + \sum_{j\in N(i)} m^t_{ji}(\bfx_i),$
where $q^t_i(\bfx_i)$ receives the message $m^t_{ji}(\bfx_i)$ from its neighboring nodes $j\in N(i)$ and the message from observed node $m^t_{\text{obs},i}(\bfx_i)$, forming a message passing algorithm for $1\leq t\leq T$. As such, the mean-field $q(\mathbf{X})$ will iteratively converge to $p(\mathbf{X}|\mathbf{o}, \Omega)$.

\subsubsection{Relational Reasoning in Mean-field GNN}
We proposed a GNN to conduct relational reasoning on the conditional random field. The GNN approximates with the mean-field message passing algorithm. \newupdate{To represent complex spatial distributions conditioned on high-dimensional linguistic features, we follow~\cite{dai2016discriminative} to map the spatial variables into high-dimensional feature spaces and learn an approximate message-passing function between these variables.} To handle multimodal output distributions better, we develop a method based on~\cite{ma2020particle} to represent the factor of mean-field $q_i$ as particles $\{(h^t_{i,k}, w^t_{i,k})\}$ in message passing, where each particle encodes the position of the corresponding object.

The GNN module takes the embeddings of spatial relations $r_{ij}=\varphi^n_\text{rel}$ as the edge attributes to compute the message.
We initialize the particles according to a normal distribution with uniform weights: $h^0_{i,k}\sim\mathcal{N}(\mathbf{0},\mathbf{I}), w_{i,k}^0=1/K$. 
In message passing, we uses a message network $f_\psi$ to output new message by the previous inputs $h^{t-1}_{j,k}$ and edge attributes $r_{ij}$: $m_{ji,k}^t=f_\psi(h_{j,k}^{t-1}, r_{ij})$. We then computes the embedding of observation by deepset~\cite{zaheer2017deep}: $m_{\text{obs},j}^t= f_\phi (\sum_{m=1}^Mb_m^jf_\text{pos}(\bfy_m))$, where $f_\phi$ and $f_\text{pos}$ are feed-forward networks. Following the form of mean-field inference, we compute the weights of the message $w_{ji, k}^t$ by the weights of particles and scores from observations computed by a weighting network $g_\phi$: $w_{ji, k}^t=\eta w_{j,k}^{t-1}\log g_\phi(h_{j,k}^{t-1}, m^t_{\text{obs},j})$. 

Then, we aggregate the message from neighboring nodes and observations to compute the node features and beliefs.
First, we sum up the messages and compute the new particles by an aggregate network $f_u$: $h_{i,k}^t=f_u(\sum_{j\in N(i)}m_{ji, k}^t + m_{\text{obs},i}^t)$. We follow the mean-field inference to evaluate new particle weights by reweighting network $g_\phi$: $\log w_{i,k}^t=c + \log g_\phi(h_{i,k}^{t}, m^t_{\text{obs}, i}) + \sum_{j\in N(i)} w_{ji,k}^{t}$. 
Next, we use resampling to avoid particle degeneracy problems. Particle degeneracy refers to the weights of all but one particle being close to zero, resulting in extremely high variance.
We resample after each iteration step to build a new set of particles from the original particles: $\{(h'^t_{i,k}, w'^t_{i,k})\}_{k=1}^K=\text{SoftResamp}(\{(h^t_{i,k}, w^t_{i,k})\}_{k=1}^K)$. However, this operation is not differentiable. 
Soft resampling~\cite{ma2020particle} makes this process differentiable via sampling from a mixture of distribution $q(j)=\alpha w_j+(1-\alpha)\frac{1}{K}$, and the new weights are evaluated using importance sampling, resulting in new belief: $w'_k=\frac{w_j}{(\alpha w_j + (1-\alpha)(1/K))}$. When $\alpha > 0$, soft resampling produces non-zero gradients for backpropagation. We take $\alpha=0.5$ in training and $\alpha=1.0$ in inference. When $t=T$, we decode the node features to get the final predictions of the target placement $\bfx_{i,k}^*=\pi_\text{dec}(h_{i,k}'^T)$ and the weights $w_{i,k}^*=w_{i,k}'^T$. 


\section{End-to-end Training}

We train the model end-to-end by minimizing two objectives. One is the negative log-likelihood of dataset $\bar{\mathcal{X}}$: $\mathcal{J}_l=\sum_{\bar{\bfx}\in\bar{\mathcal{X}}}-\log \sum_{k} w^*_k\mathcal{N}(\bar{\bfx};\mathbf{x}^*_{k}, \Sigma)$, where $\bar{\bfx}\in\bar{\mathcal{X}}$ is the labeled placement and the likelihood is a mixture of Gaussian represented by each output particle of target object $\{(\mathbf{x}^*_k, w^*_k)\}_{k=1}^K$. We also minimize the auxiliary loss for improving soft parsing: $\mathcal{J}_s=\sum_{\ell\in\bar{\mathcal{X}}}||\ell - \tilde{\ell}||^2=\sum_{\ell\in\bar{\mathcal{X}}}\sum_l (\omega_l - \tilde{\omega}_l)^2$. As such, the final objective is: $\mathcal{J}=\mathcal{J}_l + \lambda\mathcal{J}_s$, where $\lambda$ is a hyper-parameter.

%% file: 4_exp.tex
\section{Experiments}\label{sec:result}
\subsection{Experimental Setup}
\begin{figure*}
    \includegraphics[width=\textwidth]{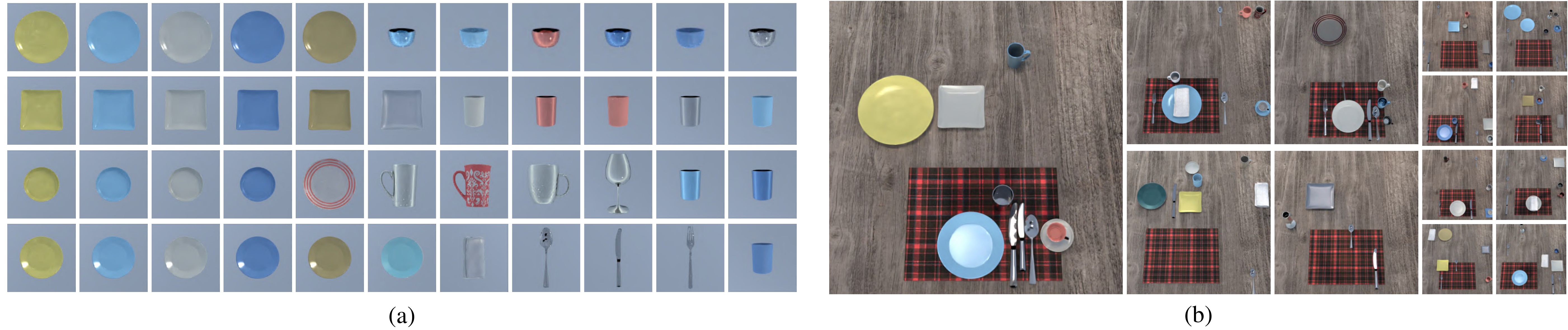}
    \caption{Images in the Tabletop dataset. The dataset contains more than 40 objects shown in (a) with randomly sampled materials and colors under random lighting conditions. The scenes shown in (b) are visually realistic and reflect the challenges in the real world.}\label{fig:dataset}
\end{figure*}

\subsubsection{Tabletop Dataset} The tabletop dataset, as shown in Fig.~\ref{fig:dataset}, consists of 30K visually realistic scenes generated by PyBullet~\cite{coumans2021} and NVISII~\cite{morrical2021nvisii}. 
The objects are sampled from 48 objects with various shapes and colors. The images contain random lighting conditions, light reflections, and partial occlusions, reflecting the challenging language grounding in the real world. We use Mask-RCNN~\cite{he2017maskrcnn} for object detection. We also select three tasks in CLIPort's benchmark: placing objects inside a bowl or a box to test object reference grounding without positional ambiguity involved. Due to the lack of bounding box labels, we use the unsupervised object detector SPACE~\cite{Lin2020SPACE} for CLIPort's tasks. Other tasks of CLIPort's dataset focus on assembly or deformable object manipulation. They are not in the scope of our research. 

We prepare \textit{human-labeled instructions} to test the model with requests using human-provided natural language.
However, collecting human-labeled data is expensive, while synthesized data is easily generated. Hence, we use synthesized \textit{structured language instructions} to train the models and fine-tune the model using human-labeled data. The training dataset of structured language instructions contains 20K instructions with single spatial relations and 20K with compositions of multiple spatial relations. Our testing dataset contains 15K instructions, including instructions with unseen compositions of seen spatial relations. We also have instructions containing ambiguous reference expressions, i.e., multiple objects are semantically identical to a reference expression.
We prepare 9K human-labeled instructions, where 7K instructions are for training, and the remaining 2K instructions are for testing. The human-labeled instructions are pre-collected from Mechanical Turk. We use image pairs to show the scene before and after an object is moved and let humans provide language requests for such object placement. 

\subsubsection{Evaluation Metric} We evaluate the success rate of object placement, repeated 5 times. The successful placements should satisfy all the spatial constraints given in the instructions. The placement should not be too far away from the reference objects with a threshold of 0.4 meters, which is approximated from the human-labeled dataset. The evaluated models are trained for 300K steps with a batch size of 1. 
We compare our performance with CLIPort~\cite{shridhar2021cliport}. \newupdate{Many other methods have restrictions in natural language (restricted language expressions) or visual scene (e.g., without referential ambiguity) and are not well-suited for our setting.} 

\begin{figure*}
    \centering
    \includegraphics[width=\textwidth]{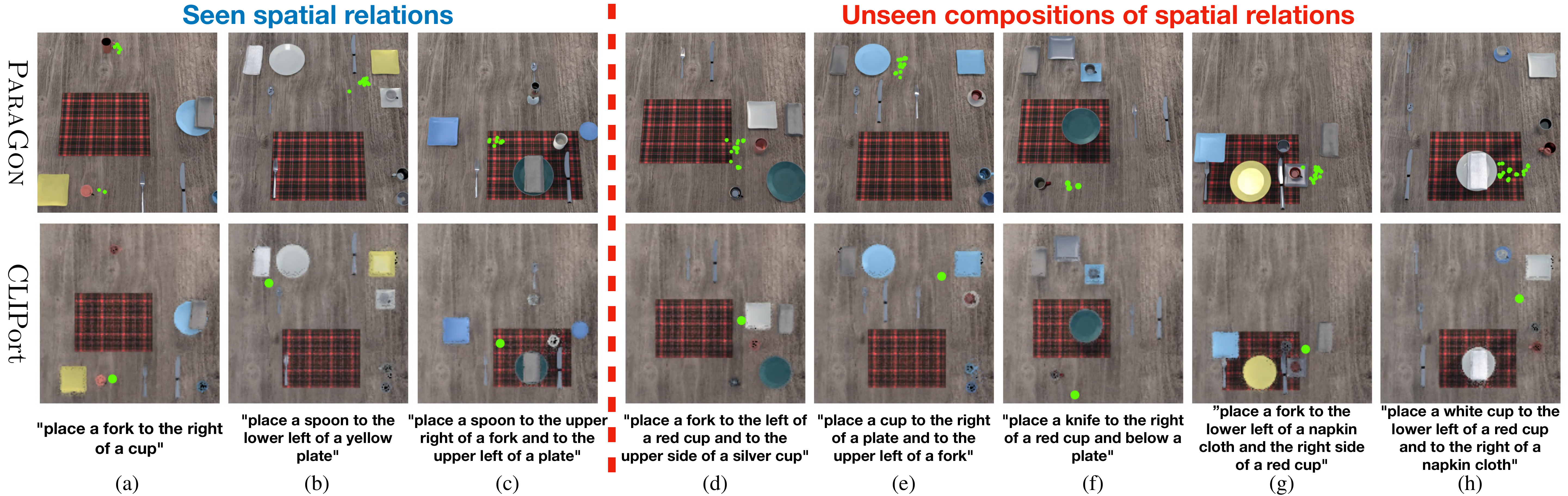}
    \caption{Demo of placement generation by \textsc{ParaGon} and CLIPort in an unseen visual scene. The green dots in the first row of images are the weighted particles of placement generated by \textsc{ParaGon}. The green dots in the second row are $\arg\max$ of placement affordance generated by CLIPort. There are demos of instructions with one spatial relation (a), seen compositions of spatial relations (b, c), unseen compositions of spatial relations (d, e, f, g, h), and ambiguous case (a). The results show that \textsc{ParaGon} can reason complex spatial relations for suitable placements and exhibits a sense of generalizability to unseen compositions. CLIPort hardly generates correct placement for unseen compositional instructions, indicating its poor compositional generalizability. }\vspace{-5pt}
    \label{fig:examples}
\end{figure*}

\begin{table}
\caption{Results on CLIPort's Dataset: Success Rate (\%)}
\centering\scriptsize
\makebox[\columnwidth][c]{
\begin{tabular}{P{12mm}P{14mm}P{22mm}P{19mm}}
\toprule
    {\scriptsize Tasks} & 
    
    \multicolumn{1}{c}{\scriptsize packing-shapes} & 
    \multicolumn{1}{c}{\scriptsize packing-google-objects} &\multicolumn{1}{c}{\scriptsize put-blocks-in-bowls} \\
    \cmidrule(lr){1-4} 
    
    \scriptsize CLIPort~\cite{shridhar2021cliport} & 99.8  & 94.6 & 99.4  \\
    \scriptsize\textsc{ParaGon}  & 98.3 & 97.3 & 99.2  \\
    \bottomrule
\end{tabular}}\vspace{-1mm}
\label{table: cliport}
\vspace{-5pt}
\end{table}

\begin{table}
    \centering\scriptsize
    \caption{Results on Tabletop Dataset: Success Rate (\%)}
    \begin{tabular}{P{8mm}P{15mm}P{4.5mm}P{4.5mm}P{4.5mm}P{4.5mm}P{4.5mm}P{4.5mm}}
        \toprule
        \multirow{2}{*}{\parbox{8mm}{\centering\% of\\Training\\Data}}& {\scriptsize Scene} & 
        \multicolumn{3}{c}{\scriptsize No Ref ambiguity} & 
        \multicolumn{3}{c}{\scriptsize With Ref Ambiguity} \\ \cmidrule(lr){2-2}\cmidrule(lr){3-5}\cmidrule(lr){6-8}
        &{\scriptsize Relation Type}&{\scriptsize{Single}}&{\scriptsize{Comp}}&{\scriptsize{Comp*}}&{\scriptsize{Single}}&{\scriptsize{Comp}}&{\scriptsize{Comp*}}\\ \cmidrule(lr){1-8} 
        \multirow{2}{*}{{100\%}} &\scriptsize CLIPort~\cite{shridhar2021cliport} & 73.2  & 69.1 & 59.5 & 71.4  & 68.2 & 51.7 \\
        &\scriptsize \textsc{ParaGon}  & \textbf{93.5}  & \textbf{92.1} &  \textbf{90.2}& \textbf{93.3}  & \textbf{91.6}  & \textbf{89.4}  \\
        \cmidrule(lr){1-8} 
        \multirow{2}{*}{{10\%}} &\scriptsize CLIPort & 57.2  & 46.3 & 36.1 & 51.0  & 44.3 & 33.2 \\
        &\scriptsize  \textsc{ParaGon} & 89.7  & 88.2 & 88.0 & 89.9  & 87.6  & 87.3 \\
        \cmidrule(lr){1-8} 
        \multirow{2}{*}{{2\%}} &\scriptsize CLIPort &  39.7 & 29.1 & 29.5 &  33.9  & 28.5 & 22.9 \\
        &\scriptsize \textsc{ParaGon} & 86.0  & 69.4 & 64.1 & 85.1  & 67.9  & 65.2\\
        \bottomrule
    \end{tabular}
    \label{table: tabletop}
    \end{table}

\subsection{Structured Language Instructions}

We first use three tasks from CLIPort's benchmark to test the models in grounding referred objects with referential ambiguity. 
We then use the Tabletop dataset to test language grounding with the presence of referential ambiguity, positional ambiguity, and compositionality. We show the results on CLIPort's benchmark and Tabletop dataset in Table~\ref{table: cliport} and Table~\ref{table: tabletop}, respectively.

\subsubsection{Object Reference Grounding}
 In CLIPort's benchmark, spatial relations are simple, as the placements are always inside a specific object. As such, learning placement generation equals learning object grounding. In these tasks, \textit{packing-shapes} requires object grounding without referential ambiguity, \textit{packing-google-objects} contains referential ambiguity for target objects, and \textit{put-blocks-in-bowls} involves referential ambiguity for context objects. Table~\ref{table: cliport} reveals that both CLIPort and \textsc{ParaGon} have good performance in object grounding even with referential ambiguity. 

\subsubsection{Object Placing with Ambiguity}
The Tabletop dataset involves placing the object next to a context object, causing positional ambiguity. Object grounding no longer equals object placement as the placement is not inside a specific object. Table~\ref{table: tabletop} suggests that \textsc{ParaGon} has a better performance with positional ambiguity. CLIPort uses convolutional architecture, which can capture strong local correlations. This inductive bias is useful if the placement is inside an object. However, positional ambiguity is only in weak agreement with the inductive bias as pixels far away can also be closely related, compromising the performance of CLIPort. \textsc{ParaGon} has an encoded graphical model for the spatial relations between objects to generate placement. It learns a distribution of the usual distance for placement from data and naturally adapts to positional ambiguity.

\begin{figure}[t]
    \centering
    \includegraphics[width=\columnwidth]{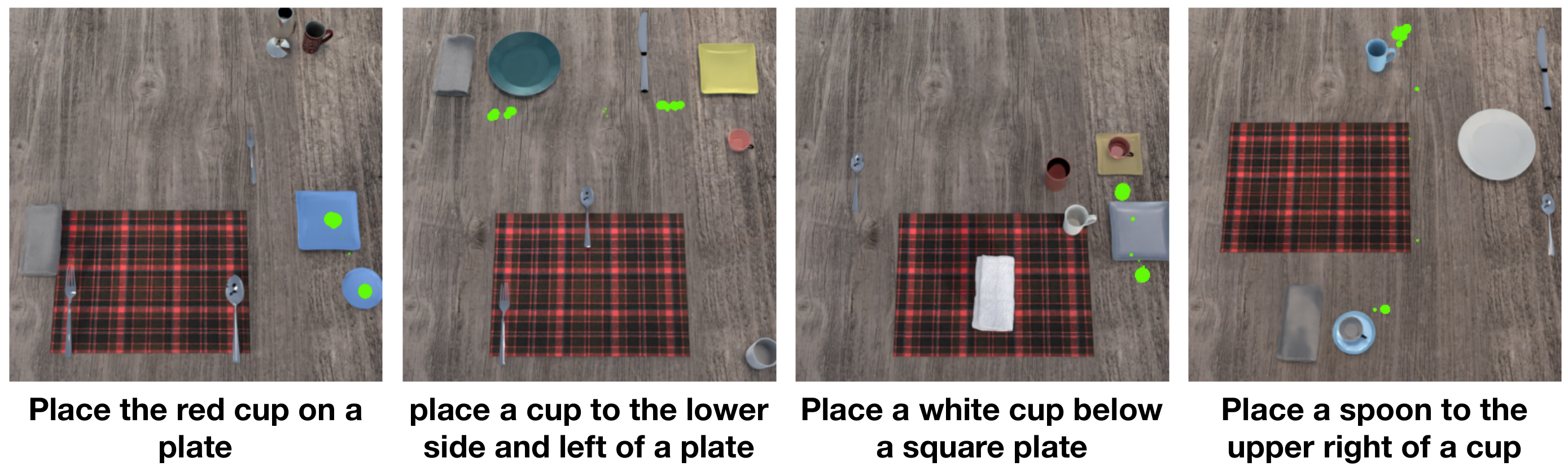}
    \caption{ Grounding instructions with referential ambiguity results in a multimodal distribution of object placement. We use particle-based GNN to capture the multimodality to adapt to referential ambiguity.}
    \label{fig: multimodal}
\end{figure}

We also test the performance of object placement with referential ambiguity (\textit{With Ref Ambiguity} in Table \ref{table: tabletop}), in which the scenes contain a few semantically identical objects. Referential ambiguity makes the correct placement non-unique, resulting in a multimodal distribution of correct placement. As shown in Table~\ref{table: tabletop}, \textsc{ParaGon} performs well with referential ambiguity. The core idea is to represent a distribution as a set of particles to capture multimodality and employ this idea in GNN for placement generation. Fig.~\ref{fig: multimodal} demonstrates that the GNN outputs a multimodal distribution when there is referential ambiguity.

\subsubsection{Compositionality}

Table~\ref{table: tabletop} also reports the results on instructions with seen and unseen compositions of seen spatial relations (\textit{Comp} and \textit{Comp*} in Table~\ref{table: tabletop}, respectively), as well as the performance of the models trained with 2\% and 10\% of the training data.
The composition substantially increases the complexity of language comprehension and the data required for training. Table~\ref{table: tabletop} shows that \textsc{ParaGon} is data-efficient in learning structured compositional instructions and can generalize to instructions with unseen compositions. It converts a complex language into a set of simple, structured relations represented as triplets to reduce complexity. Our approach operates on the grammatical structure of natural language that is generalizable to different semantic meanings. CLIPort uses single embeddings to ``memorize'' seen compositions. It has poorer generalization in the presence of composition when training data is limited and does not generalize well to unseen compositions.

\subsection{Human-labeled Instructions} 
We use human-labeled instructions to test natural human requests in noisy natural language. The instructions in this testing dataset may not be grammatically correct and contain unseen, noisy expressions for objects and spatial relations. The results are shown in Table~\ref{table:res3}, where \textit{Fine-Tuned} means the models are fine-tuned by the human-labeled dataset, and \textit{Not Fine-Tuned} models are trained only on structured instructions. The results show that \textsc{ParaGon} performs better on human-labeled instructions. \textsc{ParaGon} is data-driven and optimizes all the modules to adapt to imperfect, noisy linguistic data and extract useful relational information. \textsc{ParaGon} extracts relations from grammatical structures of instructions, which is highly generalizable and helps tackle unseen expressions. CLIPort is also data-driven but represents instructions as single embeddings that do not generalize well to unseen, noisy language expressions.  

\begin{table}
    \centering\scriptsize
    \caption{Results on Human Instructions: Success Rate (\%)
    }
    \begin{tabular}{P{20mm}P{10mm}P{10mm}P{10mm}P{10mm}}
    \toprule
        {\scriptsize } & 
        \multicolumn{2}{c}{\scriptsize Fine-Tuned} & 
        \multicolumn{2}{c}{\scriptsize Not Fine-Tuned} \\
        \cmidrule(lr){2-3}\cmidrule(lr){4-5}
        \scriptsize Method & 
        \multicolumn{1}{c}{\scriptsize \textsc{ParaGon}} & 
        \multicolumn{1}{c}{\scriptsize CLIPort} &
        \multicolumn{1}{c}{\scriptsize \textsc{ParaGon}} & 
        \multicolumn{1}{c}{\scriptsize CLIPort} \\
        \cmidrule(lr){1-5} 
        Success Rate & 81.9 & 72.5 & 70.4 & 61.3 \\
        \bottomrule
    \end{tabular}
    \label{table:res3}
    \vspace{-5pt}
    \end{table}

\subsection{Ablation Study}

\begin{table}[t]
    \centering
    \caption{Ablation Study: Success Rate (\%) 
    }\scriptsize
    \begin{tabular}{P{5mm}P{21mm}P{4.5mm}P{4mm}P{4.5mm}P{4.5mm}P{4mm}P{4.5mm}}
        \toprule
        \multirow{2}{*}{\parbox{8mm}{\centering\% of\\Training\\Data}}& {\scriptsize Scene} & 
        \multicolumn{3}{c}{\scriptsize No Ref ambiguity} & 
        \multicolumn{3}{c}{\scriptsize With Ref Ambiguity} \\ \cmidrule(lr){2-2}\cmidrule(lr){3-5}\cmidrule(lr){6-8}
        &{\scriptsize Relation Type}&{\scriptsize{Single}}&{\scriptsize{Comp}}&{\scriptsize{Comp*}}&{\scriptsize{Single}}&{\scriptsize{Comp}}&{\scriptsize{Comp*}}\\ \cmidrule(lr){1-8} 
        \multirow{5}{*}{{100\%}} & \scriptsize \textsc{ParaGon}  & \textbf{93.5}  & {92.1} &  90.2 & \textbf{93.3}  & \textbf{91.6}  & 89.4 \\
        &\scriptsize No Soft Parsing & 73.3  & 69.4 & 33.1 & 80.3 & 71.0  & 36.6 \\
        & \scriptsize No Particle  & 93.1 & \textbf{92.3} & \textbf{90.5} & 88.3  & 87.4 & 84.1 \\
        &\scriptsize No Resampling &  93.2 & 92.0 & 81.2 & 86.3 & 90.7 & 78.9 \\
        &\scriptsize \textsc{ParaGon}(ViT+Bert)  & 91.1  & 92.0 & 88.3 & 90.8  & 91.4  & \textbf{89.9} \\
        \cmidrule(lr){1-8} 
        \multirow{2}{*}{{10\%}}&\scriptsize  \textsc{ParaGon} & 89.7  & 88.2 & 88.0 & 89.9  & 87.6  & 87.3 \\
        &\scriptsize\textsc{ParaGon}(ViT+Bert)  & 88.4  & 86.0 & 85.2 & 87.7  & 86.5 & 85.4 \\
        \cmidrule(lr){1-8} 
        \multirow{2}{*}{{2\%}}&\scriptsize \textsc{ParaGon} & 86.0  & 69.4 & 64.1 & 85.1  & 67.9  & 65.2\\
        & \scriptsize\textsc{ParaGon}(ViT+Bert)  & 70.1  & 65.3 & 58.6 & 69.8  & 61.1  & 52.7 \\
        \bottomrule
    \end{tabular}
    \label{table: ablation}
\end{table}

We conduct an ablation study to assess each module's contribution in \textsc{ParaGon}. We design a variation of \textsc{ParaGon} without soft parsing, using single embeddings of language instructions to ground objects and represent relations. We also assess a version of \textsc{ParaGon} that uses a mean-field message passing neural network~\cite{dai2016discriminative} without the use of particles, and test \textsc{ParaGon} without resampling. \textsc{ParaGon} uses CLIP, a pre-trained visual-language model, to encode text and visual inputs for object reference grounding. To test the impact of using pre-trained models that are trained separately for vision and language, we replace CLIP in \textsc{ParaGon} with Visual Transformer~\cite{dosovitskiy2020image} and BERT~\cite{devlin2018bert}. 

We report ablation study results in Table \ref{table: ablation}. 
The results for \textit{No Soft Parsing} demonstrate that soft parsing is essential in learning compositional instructions. It converts complex language sentences into simple phrases for objects and relations. The embeddings of those phrases have more straightforward semantic meanings than the entire sentence and are much easier for them to be grounded in the visual scene. 
As shown in the row of \textit{No Particle}, using GNN without particles cannot capture multimodal distribution when referential ambiguity occurs and compromise the performance. 
The results in \textit{No Resampling} indicate that resampling is helpful because particle degeneracy can occur without resampling, compromising the performance of capturing multimodal distributions with referential ambiguity. 
The last 4 rows in Table \ref{table: ablation} show that \textsc{ParaGon} requires fewer data to obtain good results using pre-trained CLIP than pre-trained ViT+Bert. CLIP is pre-trained by a large dataset of image-caption pairs and is better for aligning language with visual features.


%% file: 5_conclustion.tex
\section{Conclusion}\label{sec:conclusion}

\textsc{ParaGon} integrates parsing into a probabilistic, data-driven framework for language-conditioned object placement. It combines the strength of parsing-based and embedding-based approaches for human language grounding. \textsc{ParaGon} reduces the complexity of embedding-based grounding by parsing complex sentences into simple, compositional structures and learns generalizable parsing rules from data to improve robustness. Our experiments use real, noisy human instructions and photo-realistic images, reflecting the difficulties of language grounding in realistic situations to show \textsc{ParaGon}'s capability. 
For future research, grounding 3D placement such as ``place a knife to lean against a bowl'' is interesting. As pre-trained large language models greatly impact the field through their powerful capability of content generation, leveraging large language models to parse complex instructions also helps improve our approach. 